\documentclass[11pt,a4paper]{article}
\usepackage[hyperref]{acl2020}
\usepackage{times}
\usepackage{latexsym}
\usepackage{textcomp}
\usepackage{graphicx}

\usepackage{microtype}

\aclfinalcopy 

\title{Domain Adaptation of NMT models for English-Hindi Machine Translation Task at AdapMT ICON 2020}

\author{
  Ramchandra Joshi$^{1}$, Rushabh Karnavat$^{1}$, Kaustubh Jirapure$^{1}$, Raviraj Joshi$^{2}$ \\
  $^1$Pune Institute of Computer Technology, Pune\\
  $^2$Indian Institute of Technology Madras, Chennai\\
  \texttt{\{rbjoshi1309,rpkarnavat,kaustubhmjirapure,ravirajoshi\}}@gmail.com
  }

\date{}

\begin{document}
\maketitle
\begin{abstract}
Recent advancements in Neural Machine Translation (NMT) models have proved to produce a state of the art results on machine translation for low resource Indian languages. This paper describes the neural machine translation systems for the English-Hindi language presented in AdapMT Shared Task ICON 2020. The shared task aims to build a translation system for Indian languages in specific domains like Artificial Intelligence (AI) and Chemistry using a small in-domain parallel corpus. We evaluated the effectiveness of two popular NMT models i.e, LSTM, and Transformer architectures for the English-Hindi machine translation task based on BLEU scores. We train these models primarily using the out of domain data and employ simple domain adaptation techniques based on the characteristics of the in-domain dataset. The fine-tuning and mixed-domain data approaches are used for domain adaptation. Our team was ranked first in the chemistry and general domain En-Hi translation task and second in the AI domain En-Hi translation task.

\end{abstract}

\section{Introduction}

Machine understanding of natural language queries is of paramount importance to automate different workflows. The natural language query can be in the form of text or speech. Processing of query in the form of text is more popular and easy than directly processing the raw speech waveform. The text-based Natural Language Processing (NLP) include tasks like classification, token tagging, summarization, and translation. 
Machine translation is an NLP technique to translate a sentence from a source language to a target language. The Neural Machine Translation (NMT) is a recent approach to translation producing state of the art results \cite{bahdanau2014neural}. NMT defines translation as a sequence to sequence task and uses sequence-based neural architectures like Long Short Term Memory (LSTM) \cite{hochreiter1997long} and Transformer \cite{vaswani2017attention}. Traditional techniques like rule-based translation and Statistical Machine Translation (SMT) have been outperformed by NMT models achieving significant improvements on MT tasks. In this work, we are specifically concerned with English-Hindi neural translation.


The Hindi language is one of the most popular languages in India and the fourth most spoken language in the world. Hindi is native to India and is spoken by more than 550 million total speakers worldwide. However, the number is much less as compared to global languages like English. On similar lines, the training data for the Hindi language that is publicly available for MT tasks is relatively less as compared to other highly popular languages worldwide like English, French, and German. This is important as MT tasks require a huge amount of training data to produce remarkable results using NMT models. Hindi being a relatively low resource and morphologically rich language, the amount of research in MT tasks for the Hindi language is limited \cite{philip2019baseline}. 
As Hindi is the most widely spoken language in the Indian subcontinent and the majority of content across the globe is published in English, the research in MT tasks for English-Hindi language pair becomes highly important.

Domain adaptation of translation systems to specific domains is a common practice for low resource language pairs. The adaptation is relevant as the text in different domains can vary widely \cite{luong2015stanford}. For example, social media text and the text in literary work will be quite different from style, grammar, and abbreviations perspective. The domains can be distinguished based on topics like politics, life science, news, etc, or the style of writing like formal and informal. A translation model trained on one domain may not work well on other domains. The problem is more severe in models that use word-based representation as most of the domain-specific words will be out of vocabulary \cite{sennrich2015neural}. In this work, we explore ideas for domain adaptation for English-Hindi translation on the AdapMT Shared Task ICON 2020 data sets.

The AdapMT Shared Task ICON 2020 aims to evaluate the capability of general domain machine translation for Low Resource Indian Languages. Indian languages considered in AdapMT Shared Task ICON 2020 for translation are English-Hindi, English-Telugu, and Hindi-Telugu. The shared task also focuses on Low Resource domain adaptation of machine translation systems. The adaptation is done with the use of already publicly available parallel corpora and some small in-domain parallel data for AI and Chemistry domains. The creation of a publicly available parallel corpus for low resource Indian languages is another important goal of this task.

This paper describes the system built for the English-Hindi general MT and domain adaptation tasks held under AdapMT Shared Task ICON 2020. We experimented with two popular NMT models namely attention-based LSTM encoder-decoder architecture and the Transformer architecture. For domain adaption, we explore fine-tuning and mixed domain training approaches. We show that the mixed domain training performs better than the fine-tuning based approach for the datasets used in this work.

\section{Architecture}
In this section, we describe the two popular seq2seq neural architectures for machine translation used in this work. The encoder-decoder architecture consisting of a source side encoder and a target side decoder is used for the sequence to sequence tasks \cite{sutskever2014sequence}. The encoder encodes the text in a source language into a latent representation which is consumed by the decoder to generate the text in the target language. The decoder acts like a contextual language model generating target text by attending to the source representations. The attention mechanism is thus an integral part of encoder-decoder models which allows the decoder to focus on the right context while generating the corresponding target token.
\subsection{LSTM model}
The LSTM based encoder-decoder models use stacked LSTM layers on both encoder and decoder sides. The LSTM and GRU are commonly used recurrent neural network architectures for machine translation. In this work, we use LSTM based recurrent architecture as it is shown to give slightly better results \cite{britz2017massive}. The series of stacked LSTM layers encode the source text. The hidden state of the last LSTM layer is used as the encoded output. Subsequently, the target sequence is decoded sequentially using stacked LSTM layers. The decoder also makes use of an attention mechanism to attend to the encoder's hidden state. The additive attention and dot product attention are widely used attention mechanisms \cite{bahdanau2014neural, luong2015effective}. In this work, we restrict ourselves to the use of additive attention.  

\subsection{Transformer model}
The recently introduced Transformer model has found a home in almost all NLP tasks starting with neural machine translation \cite{vaswani2017attention}. It has helped advance the state of the art in NLP and even employed for speech and vision tasks \cite{karita2019comparative, ramachandran2019stand}. The Transformer uses the self-attention mechanism as the single most important component. For the task of translation, the Transformer is used on both the encoder and decoder side. It comprises various encoders and decoders stacked over each other. The main advantage of Transformer over LSTM is the parallelism on the encoder side which helps us fully exploit the underlying hardware. The multi-headed self-attention is another architectural change that helps in providing superior results as compared to LSTM. 

On the encoder side, the input words are converted to vector embeddings and positional encoding is added to those embeddings so that the transformer gets the sense of the order or position of words. These embeddings are then passed on to the first encoder layer of the Transformer. The encoder consists of multi-head self-attention and a feed-forward neural network. The output from one encoder layer is given as input to the next encoder layer. The output of the final encoder layer is sent to the decoder.

The decoder consists of masked-multi head attention, multi-head Attention, and a feed-forward neural network. The embeddings along with positional encodings are passed on to the first layer of the decoder. The masked multi-head Attention mechanism only pays attention to the previous words. Then, it is passed through the multi-head attention mechanism attending to the encoder state and a feed-forward neural network. The output of the decoder is passed to the linear and the softmax layer where the vector scores are turned into probabilities and the word with the highest probability is chosen as output.

\section{Domain Adaptation}
While generalization is always desirable the machine learning systems are often biased towards the domain of the training data. Each domain has a different distribution and different domain data are mixed while building general systems. In very basic terms, the vocabulary of the different domains is mostly different. Table \ref{present_percentage_table} shows the percentage of in-domain tokens that are present in publicly available English-Hindi parallel training corpus. Almost 20-40\% of the tokens are specific to the target domain and not present in the general corpus. Some terms are specific to and most frequently used in a particular domain. For some words, the meaning may be different across domains. For example, \textit{"As I said that here we have one hidden layer, you can have multiple hidden layers also"} is a sentence from AI domain. Whereas, \textit{"Triacylglycerol contains three fatty acids that are esterified to the glycerol backbone"} is from the Chemistry domain. These two sentences are very specific to their domain and rarely use in real-life conversations. To interpret them in the best way we need a domain expert or subject matter expert. Similarly to build a system that works best on a particular domain, we need to make use of domain-specific data. Now because we have the same underlying language rules irrespective of the domain we can make use of out of domain data to enhance our systems if in-domain data is less. This is exactly where domain adaptation comes into the picture. 

Domain adaptation is a form of transfer learning where we adapt a general system for a specific domain. That is we tune the model to adapt to the distribution of the target domain. It has been widely studied in the context of machine translation \cite{chu2018survey}. The adaptation techniques can either be data or model-centric. The data related approaches try to exploit the monolingual corpus of the target domain \cite{domhan2017using}. A commonly used technique is to use back-translation to expand the parallel corpus of the in-domain data \cite{sennrich2015improving}. The model-based approaches also make use of monolingual corpus from the target domain to train a language model and then do a shallow or deep fusion \cite{gulcehre2015using}. There is another set of training based technique which also go into model-centric approaches. In these approaches, the model is first trained on large out of domain parallel corpus and then re-trained or fine-tuned on the small in-domain parallel corpus. There are different variations proposed in literature where the second fine-tuning is done on a mixed parallel corpus instead of only using the in-domain corpus \cite{chu2017empirical}. The concept of domain tag was introduced in \cite{sennrich2016controlling}. The model is passed the domain label along with each training sample so that it learns to distinguish between the domains. The under-represented domains are oversampled. In this work, we evaluate the domain data fine-tuning approach and mixed-data training approach. In the first approach, we train the model on general corpus followed by in-domain corpus. In the second approach, we mixed the in-domain corpus with the general data and do a single training. Since the amount of in-domain data is very less as compared to the overall general or mixed-domain data we oversample in-domain examples while training.

\section{Experimental Setup}

\subsection{Dataset Details}

In our English to Hindi machine translation experiments, we have used the publicly available IIT Bombay (IITB) English-Hindi Parallel Corpus \cite{kunchukuttan2017iit}. The training data in the IITB corpus consists of nearly 1.5M training samples. 
The IITB training data consists of sentences from the various domain. 

In addition to this, we have also used the AI and Chemistry in-domain parallel corpus provided by AdapMT Shared Task ICON 2020 organizers for training and testing the models for respective domains. 
The AI in-domain corpus contains 4872, 400, and 401 sentences in the train, validation, and test set, respectively. The Chemistry in-domain corpus contains 4984, 300, and 397 sentences in the train, validation, and test set, respectively. The data set details are described in Table \ref{statistics_table}.

\begin{table}
\centering
\begin{tabular}{ccc}
\hline \textbf{Data} & \textbf{Sentences} & \textbf{\texttildelow Tokens} \\ \hline
IIT Bombay Train & 1561840 & 19.85M / 21.4M \\
General Test & 507 & 9k / - \\
AI Train & 4872 & 77k / 83k \\
AI Dev & 400 & 6k / 6k \\
AI Test & 401 & 7k / - \\
Chemistry Train & 4984 & 125k / 139k \\
Chemistry Dev & 300 & 7k / 8k \\
Chemistry Test & 397 & 7k / - \\
\hline
\end{tabular}
\caption{\label{font-table} Statistics of the Data (En / Hi)}
\label{statistics_table}
\end{table}

\begin{table}
\centering
\begin{tabular}{cccc}
\hline \textbf{Data} & \textbf{AI} & \textbf{Chemistry} & \textbf{General}\\ \hline
Train (U) & 47 / 68 & 64 / 60 & - \\
Dev (U) & 58 / 80 & 78 / 76 & - \\
Test (U) & 59 / - & 55 / - & 56 / - \\
Train & 78 / 90 & 81 / 86 & - \\
Dev & 77 / 90 & 81 / 87 & - \\
Test & 78 / - & 77 / - & 76 / - \\
\hline
\end{tabular}
\caption{\label{font-table} Approx. \% of AdapMT domain dataset tokens (En / Hi) present in IITB Train data. Rows with a suffix 'U' indicates unique tokens, while data with no suffix indicates all tokens}
\label{present_percentage_table}
\end{table}

\subsection{Data Processing}
The individual data samples are lowercased followed by the removal of all the special characters. For training purposes, we exempted all the sentences from IITB English-Hindi Parallel Corpus with a length greater than 20 words. This was mainly done because of resource constraints to speed up training. 
After pre-processing, we train a sentence piece sub-word tokenizer to tokenize the English, as well as Hindi sentences \cite{kudo2018sentencepiece}. We train a unigram based tokenizer with a vocab size 32k \cite{kudo2018subword}. The source and target corpus of the IITB parallel corpus was used to train the individual sentence piece models. For experiments involving domain adaptation, the domain data from the train set was also included in the sentence piece training data.

\subsection{Training Details}
In this paper, we used the LSTM and Transformers based models for the English to Hindi machine translation task. For the LSTM model-based experiments, we used an attention-based encoder-decoder LSTM architecture. The encoder side of LSTM is bi-directional and the decoder side of LSTM is unidirectional with Bahdanau additive attention mechanism. The number of layers on the encoder and decoder side is set to 1 with 512 hidden units in each layer. We have used a batch size of 128 and an embedding size of 256. Adam optimizer was used as an optimizer \cite{kingma2014adam}. The subword tokenizer is used to get the subword tokens as it is known to handle the OOV problem well. 

For the Transformer model, the encoder and decoder have 6 layers each and the number of hidden layers in each layer is set to 512. The batch size was set to 128. The number of heads used is 8 with a word embedding size of 512. The optimizer used was Adam. The models were implemented in Tensorflow 2.0 and trained for a maximum of 10 epochs. The validation loss was used to pick the best epoch. The standard greedy decoding was used for all the experiments. For longer sentences, during decoding, a simple heuristic to split the data at comma was used followed by separate translations. While this approach may not be well suited to the translation as the alignment is not always monotonous, it worked decently well given the nature of the in-domain sentences.

For our experiments with LSTM and Transformer models, we first trained the models on the IITB training corpus. The models are then retrained on in-domain AI and Chemistry parallel corpus to see the improvements in the machine translation model with the inclusion of small in-domain parallel data. In the second approach, the IITB corpus is mixed with the in-domain corpus individually and, a single training is performed. The in-domain corpus in oversampled 10 times to account for a very low in-domain corpus as compared to the general corpus.

\begin{table}
\centering
\begin{tabular}{ccc}
\hline \textbf{Model} & \textbf{AI dev} & \textbf{Che dev} \\ \hline
LSTM (only IITB) & 11.54 & 8.13 \\
Transformer (only IITB) & 10.66 & 4.73 \\
LSTM (mixed) & \textbf{16.53} & \textbf{9.86} \\
Transformer (mixed) & 12.68 & 5.07 \\
LSTM (fine-tuning) & 10.62 & 5.63 \\
Transformer (fine-tuning) & 11.60 & 4.88 \\
\hline
\end{tabular}
\caption{\label{font-table} BLEU scores on in-domain dev data (model with the suffix 'only IITB' indicates that model is trained on samples from IITB train examples only, the model with the suffix 'mixed' indicates that the model is trained on data that is obtained by mixing oversampled in-domain training data with IITB training data, the model with suffix 'fine-tuning' indicates that the model is first trained on samples from IITB training data and then re-trained on in-domain corpus)}
\label{dev_result_table}
\end{table}

\begin{table}
\centering
\begin{tabular}{cccc}
\hline \textbf{Data} & \textbf{General} & \textbf{AI} & \textbf{Chemistry}\\ \hline
Test Data & 14.81 & 19.08 & 13.95 \\
\hline
\end{tabular}
\caption{\label{font-table} BLEU scores on test data as reported by AdapMT Shared Task ICON 2020 organizers}
\label{test_result_table}
\end{table}

\section{Results and Discussion}

\begin{figure}[b]
\includegraphics[scale=0.6]{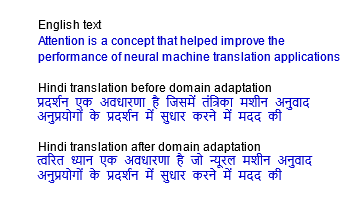}
\caption{Sample sentence from AI domain}
\label{fig:sample1}       
\end{figure}

\begin{figure}[b]
\includegraphics[scale=0.6]{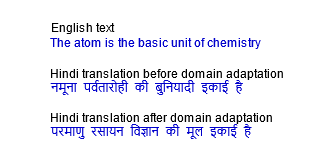}
\caption{Sample sentence from Chemistry domain}
\label{fig:sample2}       
\end{figure}

We evaluate the mixed data and fine-tuning approaches on LSTM and Transformer NMT models. To compare the models Bilingual Evaluation Understudy (BLEU) score is used \cite{papineni2002bleu}. We report the BLEU score on validation data of AI and Chemistry in-domain corpus. Table \ref{dev_result_table} shows the results for de-tokenized validation data. The mixed data training approach performs the best in comparison to the no-domain data and fine-tuning approach. The no-domain data approach performs better as compared to the fine-tuning approach. This indicates that the simple fine-tuning approach is not suited to the very small in-domain corpus and is susceptible to catastrophic forgetting. We see that although the Transformer based models perform well on the IITB test data they do not generalize well on the domain tasks. However, we feel that the low numbers with the Transformer can be enhanced using appropriate hyper-parameters and modifying the training approach. The system submitted for evaluation was LSTM based model trained on a mixed corpus which was giving the best validation scores. The results of the test system are shown in Table \ref{test_result_table}. The translations for the general test set were generated using the LSTM model trained only on the IITB parallel corpus. Some sample example translations using no-domain data and mixed-training approach is shown in Figure \ref{fig:sample1} and Figure \ref{fig:sample2}.

\section{Conclusion}

In this paper, we evaluated the effectiveness of attention-based encoder-decoder LSTM and Transformer models on a low resource English to Hindi Translation Task held under AdapMT Shared Task ICON 2020. Our experiments showed that mixed domain training works well as compared to the fine-tuning approach for domain adaptation. The addition of small in-domain parallel data can indeed improve the results on AI and Chemistry domains provided in the shared task.

\section*{Acknowledgements} This work was done under the L3Cube Pune mentorship program. We would like to thank L3Cube and our mentors for the end to end guidance and encouragement to participate in the shared task.

\bibliographystyle{acl_natbib}
\bibliography{main}

\begin{thebibliography}{22}
\expandafter\ifx\csname natexlab\endcsname\relax\def\natexlab#1{#1}\fi

\bibitem[{Bahdanau et~al.(2014)Bahdanau, Cho, and Bengio}]{bahdanau2014neural}
Dzmitry Bahdanau, Kyunghyun Cho, and Yoshua Bengio. 2014.
\newblock Neural machine translation by jointly learning to align and
  translate.
\newblock \emph{arXiv preprint arXiv:1409.0473}.

\bibitem[{Britz et~al.(2017)Britz, Goldie, Luong, and Le}]{britz2017massive}
Denny Britz, Anna Goldie, Minh-Thang Luong, and Quoc Le. 2017.
\newblock Massive exploration of neural machine translation architectures.
\newblock \emph{arXiv preprint arXiv:1703.03906}.

\bibitem[{Chu et~al.(2017)Chu, Dabre, and Kurohashi}]{chu2017empirical}
Chenhui Chu, Raj Dabre, and Sadao Kurohashi. 2017.
\newblock An empirical comparison of simple domain adaptation methods for
  neural machine translation.
\newblock \emph{arXiv preprint arXiv:1701.03214}.

\bibitem[{Chu and Wang(2018)}]{chu2018survey}
Chenhui Chu and Rui Wang. 2018.
\newblock A survey of domain adaptation for neural machine translation.
\newblock \emph{arXiv preprint arXiv:1806.00258}.

\bibitem[{Domhan and Hieber(2017)}]{domhan2017using}
Tobias Domhan and Felix Hieber. 2017.
\newblock Using target-side monolingual data for neural machine translation
  through multi-task learning.
\newblock In \emph{Proceedings of the 2017 Conference on Empirical Methods in
  Natural Language Processing}, pages 1500--1505.

\bibitem[{Gulcehre et~al.(2015)Gulcehre, Firat, Xu, Cho, Barrault, Lin,
  Bougares, Schwenk, and Bengio}]{gulcehre2015using}
Caglar Gulcehre, Orhan Firat, Kelvin Xu, Kyunghyun Cho, Loic Barrault, Huei-Chi
  Lin, Fethi Bougares, Holger Schwenk, and Yoshua Bengio. 2015.
\newblock On using monolingual corpora in neural machine translation.
\newblock \emph{arXiv preprint arXiv:1503.03535}.

\bibitem[{Hochreiter and Schmidhuber(1997)}]{hochreiter1997long}
Sepp Hochreiter and J{\"u}rgen Schmidhuber. 1997.
\newblock Long short-term memory.
\newblock \emph{Neural computation}, 9(8):1735--1780.

\bibitem[{Karita et~al.(2019)Karita, Chen, Hayashi, Hori, Inaguma, Jiang,
  Someki, Soplin, Yamamoto, Wang et~al.}]{karita2019comparative}
Shigeki Karita, Nanxin Chen, Tomoki Hayashi, Takaaki Hori, Hirofumi Inaguma,
  Ziyan Jiang, Masao Someki, Nelson Enrique~Yalta Soplin, Ryuichi Yamamoto,
  Xiaofei Wang, et~al. 2019.
\newblock A comparative study on transformer vs rnn in speech applications.
\newblock In \emph{2019 IEEE Automatic Speech Recognition and Understanding
  Workshop (ASRU)}, pages 449--456. IEEE.

\bibitem[{Kingma and Ba(2014)}]{kingma2014adam}
Diederik~P Kingma and Jimmy Ba. 2014.
\newblock Adam: A method for stochastic optimization.
\newblock \emph{arXiv preprint arXiv:1412.6980}.

\bibitem[{Kudo(2018)}]{kudo2018subword}
Taku Kudo. 2018.
\newblock Subword regularization: Improving neural network translation models
  with multiple subword candidates.
\newblock \emph{arXiv preprint arXiv:1804.10959}.

\bibitem[{Kudo and Richardson(2018)}]{kudo2018sentencepiece}
Taku Kudo and John Richardson. 2018.
\newblock Sentencepiece: A simple and language independent subword tokenizer
  and detokenizer for neural text processing.
\newblock \emph{arXiv preprint arXiv:1808.06226}.

\bibitem[{Kunchukuttan et~al.(2017)Kunchukuttan, Mehta, and
  Bhattacharyya}]{kunchukuttan2017iit}
Anoop Kunchukuttan, Pratik Mehta, and Pushpak Bhattacharyya. 2017.
\newblock The iit bombay english-hindi parallel corpus.
\newblock \emph{arXiv preprint arXiv:1710.02855}.

\bibitem[{Luong and Manning(2015)}]{luong2015stanford}
Minh-Thang Luong and Christopher~D Manning. 2015.
\newblock Stanford neural machine translation systems for spoken language
  domains.
\newblock In \emph{Proceedings of the International Workshop on Spoken Language
  Translation}, pages 76--79.

\bibitem[{Luong et~al.(2015)Luong, Pham, and Manning}]{luong2015effective}
Minh-Thang Luong, Hieu Pham, and Christopher~D Manning. 2015.
\newblock Effective approaches to attention-based neural machine translation.
\newblock \emph{arXiv preprint arXiv:1508.04025}.

\bibitem[{Papineni et~al.(2002)Papineni, Roukos, Ward, and
  Zhu}]{papineni2002bleu}
Kishore Papineni, Salim Roukos, Todd Ward, and Wei-Jing Zhu. 2002.
\newblock Bleu: a method for automatic evaluation of machine translation.
\newblock In \emph{Proceedings of the 40th annual meeting of the Association
  for Computational Linguistics}, pages 311--318.

\bibitem[{Philip et~al.(2019)Philip, Namboodiri, and
  Jawahar}]{philip2019baseline}
Jerin Philip, Vinay~P Namboodiri, and CV~Jawahar. 2019.
\newblock A baseline neural machine translation system for indian languages.
\newblock \emph{arXiv preprint arXiv:1907.12437}.

\bibitem[{Ramachandran et~al.(2019)Ramachandran, Parmar, Vaswani, Bello,
  Levskaya, and Shlens}]{ramachandran2019stand}
Prajit Ramachandran, Niki Parmar, Ashish Vaswani, Irwan Bello, Anselm Levskaya,
  and Jonathon Shlens. 2019.
\newblock Stand-alone self-attention in vision models.
\newblock \emph{arXiv preprint arXiv:1906.05909}.

\bibitem[{Sennrich et~al.(2015{\natexlab{a}})Sennrich, Haddow, and
  Birch}]{sennrich2015improving}
Rico Sennrich, Barry Haddow, and Alexandra Birch. 2015{\natexlab{a}}.
\newblock Improving neural machine translation models with monolingual data.
\newblock \emph{arXiv preprint arXiv:1511.06709}.

\bibitem[{Sennrich et~al.(2015{\natexlab{b}})Sennrich, Haddow, and
  Birch}]{sennrich2015neural}
Rico Sennrich, Barry Haddow, and Alexandra Birch. 2015{\natexlab{b}}.
\newblock Neural machine translation of rare words with subword units.
\newblock \emph{arXiv preprint arXiv:1508.07909}.

\bibitem[{Sennrich et~al.(2016)Sennrich, Haddow, and
  Birch}]{sennrich2016controlling}
Rico Sennrich, Barry Haddow, and Alexandra Birch. 2016.
\newblock Controlling politeness in neural machine translation via side
  constraints.
\newblock In \emph{Proceedings of the 2016 Conference of the North American
  Chapter of the Association for Computational Linguistics: Human Language
  Technologies}, pages 35--40.

\bibitem[{Sutskever et~al.(2014)Sutskever, Vinyals, and
  Le}]{sutskever2014sequence}
Ilya Sutskever, Oriol Vinyals, and Quoc~V Le. 2014.
\newblock Sequence to sequence learning with neural networks.
\newblock In \emph{Advances in neural information processing systems}, pages
  3104--3112.

\bibitem[{Vaswani et~al.(2017)Vaswani, Shazeer, Parmar, Uszkoreit, Jones,
  Gomez, Kaiser, and Polosukhin}]{vaswani2017attention}
Ashish Vaswani, Noam Shazeer, Niki Parmar, Jakob Uszkoreit, Llion Jones,
  Aidan~N Gomez, {\L}ukasz Kaiser, and Illia Polosukhin. 2017.
\newblock Attention is all you need.
\newblock In \emph{Advances in neural information processing systems}, pages
  5998--6008.

\end{thebibliography}

\end{document}